# Hybrid Quantum-Classical Neural Network for Incident Detection

**Zadid Khan, PhD**
Department of Civil Engineering
Clemson University, Clemson, SC 29631, USA
Email: mdzadik@clemson.edu

**Sakib Mahmud Khan, PhD**
Department of Civil Engineering
Clemson University, Clemson, SC 29631, USA
Email: sakibk@clemson.edu

**Jean Michel Tine**
Comp. Sci., Phy., and Engineering Department, Benedict College
1600 Harden Street , Columbia, SC 29204
Email: Jean-Michel.Tine82@my.benedict.edu

**Ayse Turhan Comert**
Comp. Sci., Phy., and Engineering Department, Benedict College
1600 Harden Street , Columbia, SC 29204
Email: Ayse.Turhan-Comert45@my.benedict.edu

**Diamon Rice**
Comp. Sci., Phy., and Engineering Department, Benedict College, 1600 Harden Street
Columbia, SC 29204
Email: Diamon.Rice59@my.benedict.edu

**Gurcan Comert, PhD**
Comp. Sci., Phy., and Engineering Department, Benedict College
1600 Harden Street , Columbia, SC 29204
Email: Gurcan.Comert@Benedict.edu

**Dimitra Michalaka, PhD**
Department of Civil and Environmental Engineering
The Citadel, Charleston, SC 29409, SC
Email: Dimitra.Michalaka@citadel.edu

**Judith Mwakalonge, PhD**
Department of Civil and Mechanical Engineering Technology and Nuclear Engineering
South Carolina State University
Email: jmwakalo@scsu.edu

**Reek Majumdar**
Glenn Department of Civil Engineering
Clemson University, Clemson, SC 29631, USA
Email: rmajumd@clemson.edu

**Mashrur Chowdhury, PhD**
Eugene Douglas Mays Chair in Transportation, Glenn Department of Civil Engineering
Clemson University, Clemson, SC 29631, USA
Email: mac@clemson.edu

**ABSTRACT**

The efficiency and reliability of real-time incident detection models directly impact the affected corridors' traffic safety and operational conditions. The recent emergence of cloud-based quantum computing infrastructure and innovations in noisy intermediate-scale quantum devices have revealed a new era of quantum-enhanced algorithms that can be leveraged to improve real-time incident detection accuracy. In this research, a hybrid machine learning model, which includes classical and quantum machine learning (ML) models, is developed to identify incidents using the connected vehicle (CV) data. The incident detection performance of the hybrid model is evaluated against baseline classical ML models. The framework is evaluated using data from a microsimulation tool for different incident scenarios. The results indicate that a hybrid neural network containing a 4-qubit quantum layer outperforms all other baseline models when there is a lack of training data. We have created three datasets; DS-1 with sufficient training data, and DS-2 and DS-3 with insufficient training data. The hybrid model achieves a recall of 98.9%, 98.3%, and 96.6% for DS-1, DS-2, and DS-3, respectively. For DS-2 and DS-3, the average improvement in F2-score (measures model's performance to correctly identify incidents) achieved by the hybrid model is 1.9% and 7.8%, respectively, compared to the classical models. It shows that with insufficient data, which may be common for CVs, the hybrid ML model will perform better than the classical models. With the continuing improvements of quantum computing infrastructure, the quantum ML models could be a promising alternative for CV-related applications when the available data is insufficient.

**INTRODUCTION**

Efficiency and reliability need to be ensured while detecting incidents to timely initiate incident response to restore the desirable traffic operations in any roadway corridor. Incident detection for traffic roadways is one of the most heavily investigated problems using fundamental traffic flow diagrams, statistical, rule-based, and, more recently, machine learning models. Note that the real-time incident detection accuracy of the models depends on the available data. Connected vehicles or CVs can provide a plethora of real-time data, and having data from a limited number of CVs can be used to detect incidents (*1*). In a data-driven and connected world, CVs will generate Basic Safety Message or BSM data (*2*), which will be collected by roadside units or RSUs in real-time using the available communication options. RSU, the roadside edge device, will collect data from CVs within its coverage area, and multiple RSUs can be used to create a continuous and seamless coverage along a long stretch of a freeway. Data collected from the RSUs can be shared with a cloud-based traffic management system. As shown in Figure 1, once an incident occurs and BSM data are available from CVs, the incident detection algorithm runs in the cloud server, and the information is shared with all approaching CVs.

The widespread and successful demonstration of machine learning models and emerging CV technology have motivated the authors to use data-driven models to monitor incidents. However, machine-learning models can sometimes create false negative information due to the dynamic nature of incident occurrences. Developing a more accurate incident detection system is always a target for transportation researchers. With the emergence of innovative quantum computing infrastructure and noisy intermediate-scale quantum devices, a series of quantum-only and hybrid (quantum and classical) models are now available to provide better incident detection accuracy. One such example of quantum computing-induced improvement is discussed in (*3*), where the authors have studied how to use the quantum-enhanced feature for a quantum machine learning model. Also, by controlling entanglement and interference, the quantum search space can be used for the machine learning models. This quantum search space is an exponentially large space compared to that of any classical computer.

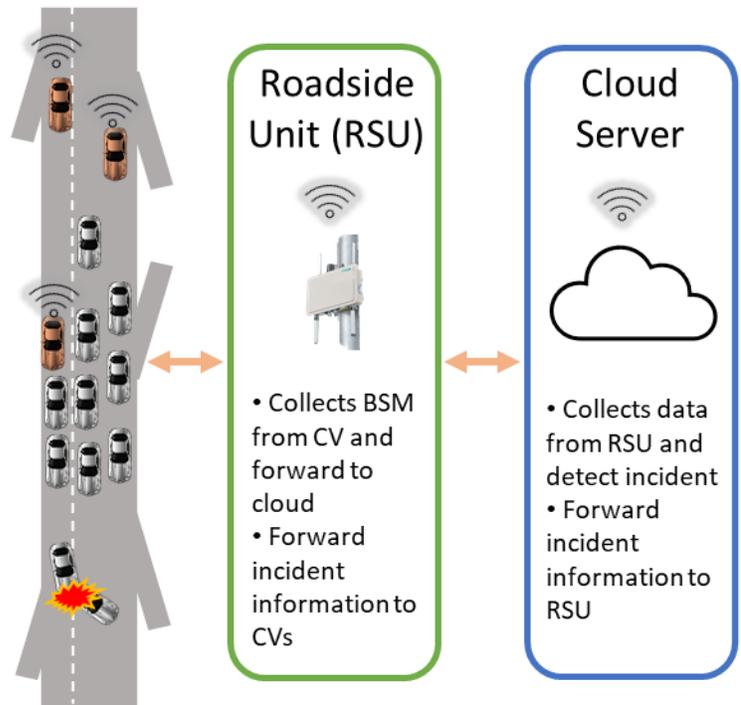

**Figure 1 Real-time incident detection with hybrid model**

This research aims to develop and evaluate a hybrid neural network model, which includes classical and quantum neural networks, for traffic incident detection. In the hybrid model, a quantum layer is placed between two classical layers to improve the neural network's performance in detecting incidents. The hybrid model will be evaluated against a classical only neural network to measure the improvement due to the use of the hybrid model. A case study is conducted in this research using a microsimulation tool (Simulation Of Urban Mobility or SUMO). With the microscopic simulation tool, multiple incident scenarios from a freeway section are simulated, and CV BSM data have been generated from a sample of the total simulated vehicles. Using these BSM data, the hybrid model is trained and later evaluated for different incident scenarios.

*Khan et al.***LITERATURE REVIEW**
This section discusses the recent development of quantum machine learning models and earlier studies where classical machine learning models were used to detect an incident. To the best of the authors' knowledge, quantum machine learning models have not been applied for incident detection.

**Quantum Machine Learning**

Quantum deep learning gained interest in predicting atomic energies and chemical potentials (*4*). Van Nieuwenburg et al. used quantum AI to learn phase changes of matter (*5*), and Pang et al. used it for learning chromodynamics (*6*). Patel et al. applied a Q-neural network (Q-NN) for signature verification and compared it against classical machine learning algorithms (*7*). Q-NN achieved 95% accuracy, where the classical neural network (NN) model achieved 89% accuracy. Patel and Tiwari (*8*) utilized Quantum Binary NN (Q-BNN) model for breast cancer classification and compared it against Gaussian Processes, NNs, multilayer perceptron, support vector machine (SVM), etc. Q-BNN was able to achieve above 95% accuracy, where other methods were all less than 80% accurate. Li et al. used quantum-behaved particle swarm optimization with binary encoding and applied it to MNIST data (*9*). The authors found that quantum-based methods provided higher accuracy compared to SVM. Chen et al. applied quantum Convolutional Neural network (CNN) for image classification and reported higher accuracy (94%) than classical CNN (90%) (*10*). Wang et al. showed that quantum stochastic networks (Q-SNN) can achieve better performance against classical networks classifying sentences (*11*). Q-SNN was able to converge faster and to higher accuracy compared to classical SNN. No study has been conducted to use quantum machine learning for incident detection.

**Traffic Incident Detection using Machine Learning**

Based on our review, we have found that using machine learning models to improve incident detection accuracy is an active field of research. For example, Karim and Adeli compared the performance of the wavelet energy algorithm and California algorithm 8 for the various rural and urban freeways (*12*). They found out wavelet algorithm performed well-detecting incidents at average 95% accuracy under 2 minutes for urban and under 3 minutes for rural freeways. Performance was only low at 59% detection for rural two-way freeways for the 10-minute incident. Sheu (*13*) used fuzzy clustering-based to detect incidents that were 27 types of 10-minute incidents. The author showed 100% accuracy in detecting incidents with less than a minute time to detect. The study was based on simulation data generated from a short 2-km 3-lane segment. Zhang and Taylor (*14*) proposed a data preparation and Bayesian Network-based incident detection framework that is adaptive to threshold values. The results showed a 92% detection rate with less than 1% false positives. The methods showed better results compared to MLF. Fangming and Han (*15*) applied BPNN using simulated data and achieved a 94% detection rate, 5% false positives, and less than a minute detection time. Zheng et al. showed SVM's detection performance in an imbalanced dataset and obtained close to 94% detection rate, 2% false positives, and less than 3-minute detection time (*16*). However, the authors did not compare the performance of SVM against other methods. Wang et al. (*17*) combined time-series and SVM to detect incidents. The study reported an 80% detection rate with 3% false positives and less than 10 minutes of detection.

Shang et al. (*18*) compared machine learning methods for incident detection. The authors' hybrid method achieved a 96% detection rate, 2% false positives, and less than 3 minutes detection time. Ensemble methods provided comparable above 90% detection rates, and long-short term memory (LSTM) resulted in less than 70% detection. Jiang and Deng (*19*) proposed factor analysis and weighted random forest and obtained a 98% incident detection rate with 1% false positives. Fang et al. (*20*) applied deep learning methods. According to the study, deep learning models with the variable selection provided above 95% detection rate, less than 1% false positives, and less than a minute detection time. Chakraborty et al. (*21*) used semi-supervised learning to address trajectory classification for incident detection and achieved above

*Khan et al.*

80% detection rate using video data. Yang et al. (*22*) applied an autoencoder and obtained a 90% incident detection rate, 4% false positives, and 4-minute time to detection.

**METHOD**

In this research, an incident detection framework is developed that relies on real-time CV data, which include average CV speed and CV count. Using a simulation study, the framework is validated, and the classical only and hybrid Q-AI models have been evaluated.

**Incident Detection Framework**

The incident detection framework relies on CV movement in specific corridor segments, which are referred to as zones. Zones are the evenly distributed spatial segments along the corridor (as shown in Figure 2). Each RSU identifies the operational condition of each zone based on the CV data from that zone. Data are aggregated for each second.

The incident detection steps are shown in Figure 3. The first stage is collecting real-time connected vehicle BSM data, containing CV positions, CV speed, etc. After that, the data is aggregated per zone. Therefore, it gives a zone-wise average value of CV speed and CV count. Six features are created for the target zone, and its upstream and downstream zones are created, and the corresponding values are extracted. These six features are the input to the incident detection model. The six features are average speed and count of that zone, average speed and count of the upstream zone, and average speed and count of the downstream zone. These six features are input the hybrid neural network containing classical and quantum layers. The output of the model is a binary classification value, indicating if there has been any incident in the zone of interest or not.

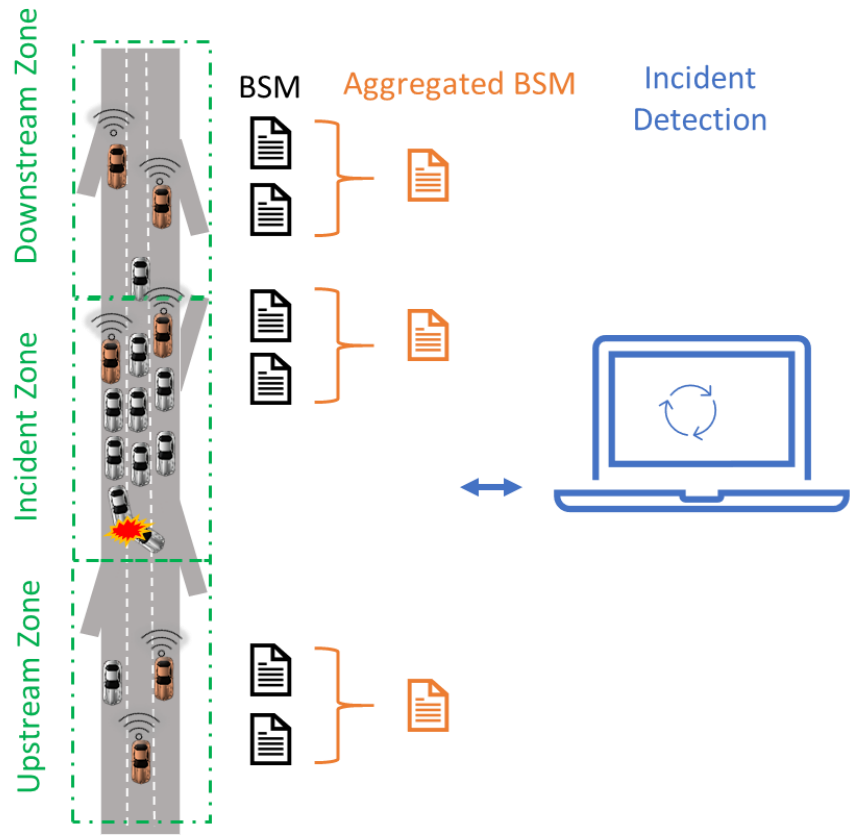

**Figure 2 Incident detection using aggregated CV data**

To develop the hybrid model, a quantum layer is used between classical fully connected layers of a neural network. The quantum layer is developed using a quantum node, where a quantum function can be operated using quantum operators such as gates, quantum state preparations, measurements, and noisy channels. Using backpropagation, the parameters of the hybrid model can be optimized on a simulated or real-world quantum device. Finally, the observations from the quantum layers are measured and fed into another classical layer. This is how we create a hybrid model that contains both types of layers.

*Khan et al.*

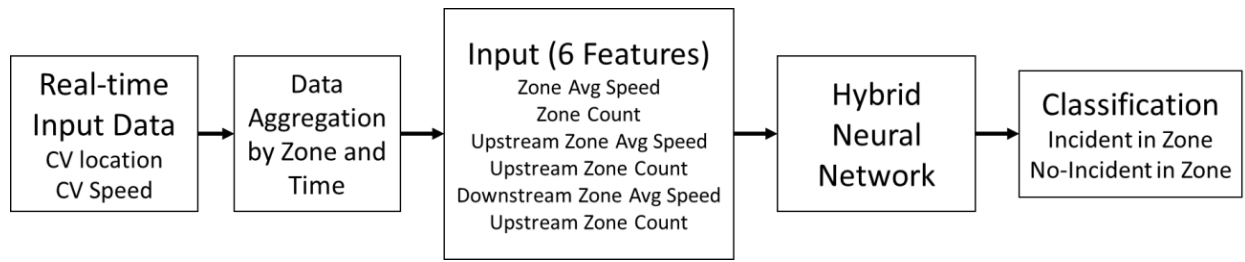

**Figure 3 Incident detection framework**

**Case Study**

*Simulation Model Creation*

In this study, we have designed a portion of the I-85 interstate between Greenville, SC, and Atlanta, GA, in the USA. We have gathered the road network data from Openstreetmap. Openstreetmap generates an OSM file containing all the information. We have used SUMO to simulate the interstate network. Netconvert and Netedit are used to design the road network appropriately. The interstate network contains three interchanges. Each interchange contains on-ramps and off-ramps for vehicles exiting and entering the interstate. It has two lanes in the northbound direction and two lanes in the southbound direction. The arterials also have two lanes. The ramps are single lane. Figure 4 shows the road network considered in this study. Figure 5 shows normal traffic operations on the interstate, and Figure 6 shows normal traffic operations in the interchanges and ramps.

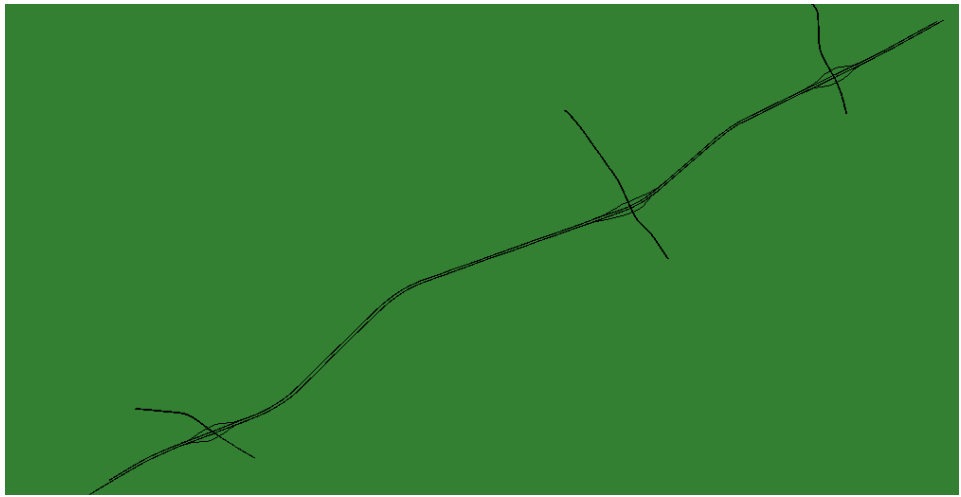

**Figure 4 I-85 road network in SUMO**

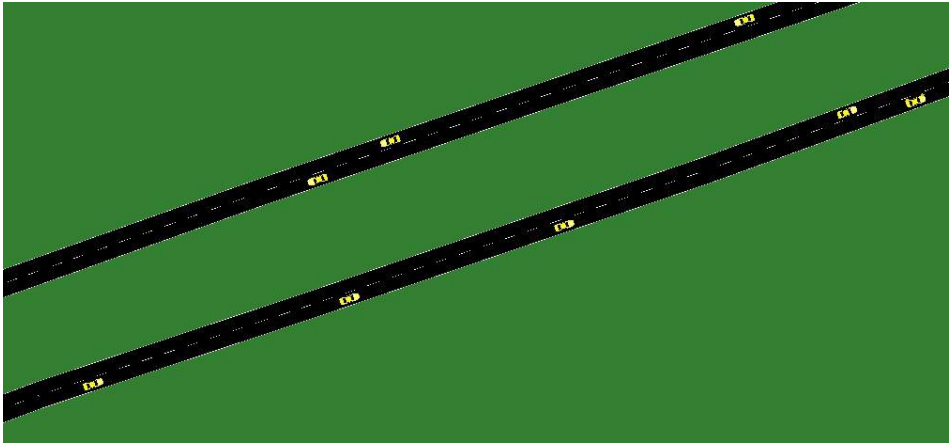

**Figure 5 I-85 interstate normal traffic condition**

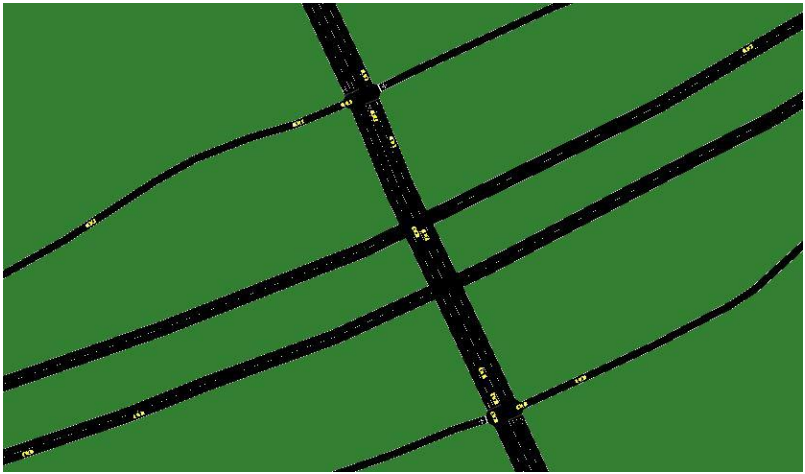

**Figure 6 I-85 interchange and ramp normal traffic condition**

After creating the network and generating traffic in SUMO, we created a method to aggregate the data by zone and collect zone-specific data. We divided the network into 56 zones, with 28 on the northbound and 28 on the southbound direction. The purpose of the simulation is to create a training dataset for the incident detection model and test the model performance. Therefore, we first need to create incidents in the simulation. In this study, we have simulated incidents by scheduling pairs of vehicles to halt on the roadway and block interstate lanes at a different time during the simulation. When an accident happens on the road, usually some vehicles are stationary on the road, and other vehicles are unable to go through the lane due to the stationary vehicles blocking their path. In this study, we have simulated this event by forcefully stopping pairs of vehicles for some time. After the incident duration has passed, the vehicles are able to start moving again. To show the effect of this scheduling in simulation, we are showing Figures 7, 8, and 9. In Figure 7, two vehicles were halted forcefully using scheduling. In Figure 8, we see the effect in the upstream and the downstream. In the downstream area, a queue has built up due to the incident. In the upstream, there are no vehicles and no traffic flow. In Figure 9, we see that the downstream vehicles have started moving.

*Khan et al.*

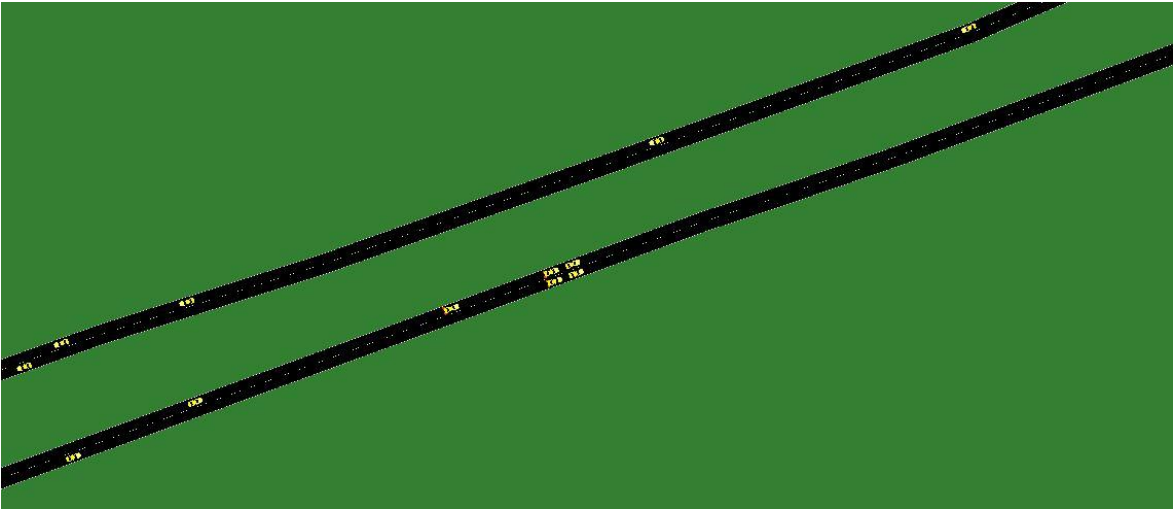

**Figure 7 An incident happening on I-85 northbound**

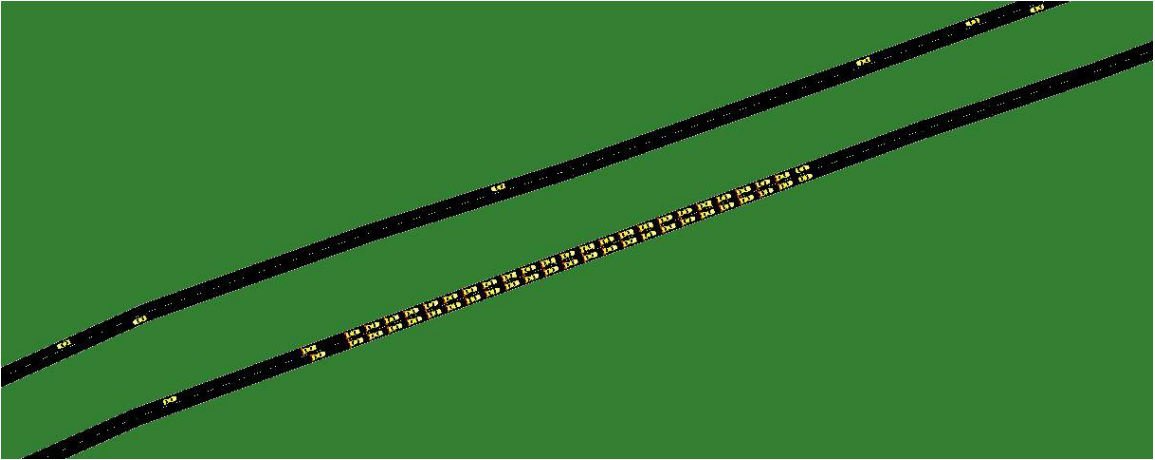

**Figure 8 Incident causing a queue buildup in the upstream direction**

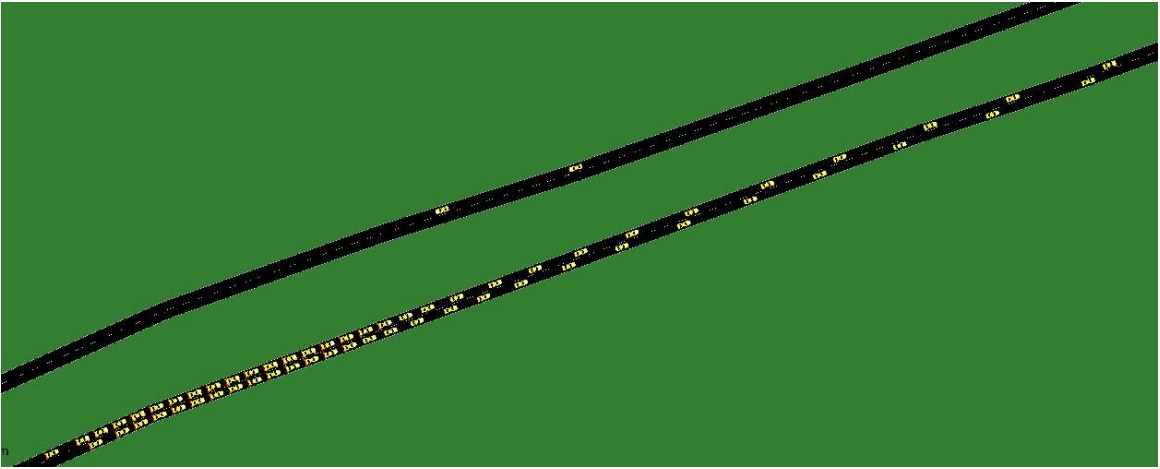

**Figure 9 Vehicles flow resuming after the incident**

*Data Generation for the Incident Detection Models*

After creating incidents in the simulation, we run the simulation and collect the raw CV data from the simulation. In this study, we have assumed 100% CV penetration. The data is aggregated by zone at first. This dataset contains zone-wise speed and vehicle count per second. Then we create the feature vectors by combining the data from different zones. We only consider the zone's data, the upstream zone data, and the downstream zone data for each dataset. This creates a dataset of 70000 rows. However, this dataset is unlabeled. As we have created the scheduling of the incidents, we create a label feature and set the corresponding rows to 1, indicating the occurrence of incidents. This dataset is split into training and testing in two ways. At first, we take 40000 rows in the training set and 30000 rows in the testing set, and we call it DS-1. Next, we take 15000 rows in the training set and 55000 rows in the testing set. We call it DS-2. The reason for creating DS-2 is, we want to evaluate the performance of a hybrid neural network when the available data is scarce. Quantum neural network should perform well when there is a lack of training data for the model. We also create a third dataset, DS-3, by aggregating by time. In DS-1 and DS-2, we aggregate the data per second. In DS-3, we aggregate the data per minute. This gives us a shortened dataset containing only 1400 rows. We split the dataset again into training and testing, and the training set only has 150 rows while the test set contains 1250 rows. This is done intentionally to test the machine learning models for extreme cases where there is a lack of data.

*Incident Detection Models*

The model architecture is shown in Figure 10. The input layer contains six neurons corresponding to the six inputs discussed previously. The next three layers are dense classical layers, containing 48, 32, and 4 neurons, respectively. After that, we add a 4-qubit quantum layer. The quantum layer takes the values from the 4-neuron classical layer as input and converts them to qubit values. The quantum circuit performs angle embedding and basic entangling. After that, these values need to be converted into classical values again, so we perform measurements to get classical values. 4 values are extracted from 4 qubits, which are again input to the 4-neuron classical layer. The final layer is the output layer containing only one neuron, which gives a binary output corresponding to incident or no-incident. All the classical layers have rectified linear unit (ReLu) activation, except the output layer, which has a sigmoid activation. The model uses an Adam optimizer, and the loss function is binary cross-entropy loss. The number of epochs is 20, and the batch size is 16. The data is normalized between 0 and 1 before it is fed to the model. The model is trained on three separate datasets, DS-1, DS-2, and DS-3. The training accuracy achieved for the 3 datasets are 98%, 99%, and 99%, respectively.

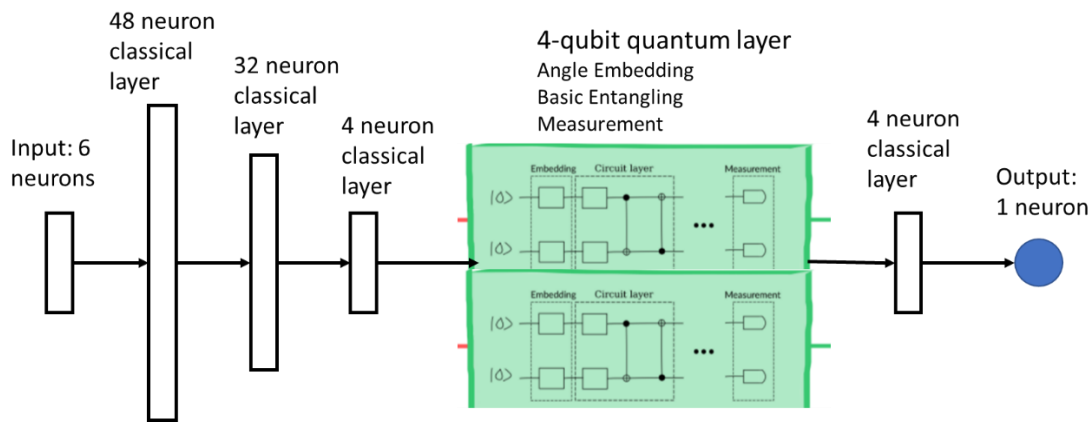

**Figure 10 Hybrid (Classical + Quantum) Neural Network for Incident Detection**

*Khan et al.*

**DATA ANALYSIS**
  In this section, we will compare the performance of the hybrid (classical+quantum) neural network with other baseline models. At first, we will discuss the implementation of baseline models and the hybrid model. Then, we will talk about evaluation metrics.
  As this is a binary classification task, we have searched the literature and identified the baseline models. We have used random forest (RF), support vector machine (SVM), extreme gradient boosting (XGB), and classical neural network (NN) as baseline models. We have implemented the RF and SVM models using the scikit-learn package in Python. For SVM, we have created two variations of the model, one with radial basis kernel, and the other with $2^{nd}$ degree polynomial kernel. For RF, we have used 100 estimators. We have used classical layers of 48 neurons, 32 neurons, and 1 neuron for the classical neural network. The classical NN is implemented using Keras and Pytorch. For the quantum layer, we have used the Pennylane package in Python to simulate a quantum device in a classical computer. In Pennylane, qubit nodes can be created using random number generators that mimic quantum qubits' behavior. Then, it also allows the creation of quantum layers, which can be readily integrated with classical Keras layers. This allows for smooth model creation and generation.
  For evaluation metrics, we are more concerned with the positive predictions, which indicate the presence of incidents. The metrics we have used are true positives, false positives, false negatives, accuracy, precision, recall, and F2-score (*23*). All models are run 30 times, and an average of the metrics values are collected.

**Results for DS-1**
  At first, we check how the models perform for the original DS-1 dataset. As mentioned in the case study section, DS-1 is a per-second zone aggregated dataset containing 70000 rows of data, 40000 rows in training, and 30000 rows in testing. For Table I, we see that all the models perform very well in this scenario. This suggests that with sufficient training data, any ML model can be trained and optimized to detect incidents accurately. Overall, the hybrid model achieves an average recall of 98.9% and an F2-score of 98.7%. So, it is very suitable for incident detection when there is sufficient training data. Since the performance of other baseline models is very similar in terms of precision, recall, and F2-score, we are not identifying any specific model in this section. All models, including the hybrid model, have good detection accuracy if there is sufficient training data.

**TABLE 1 Comparison of Model Performance for DS-1**

| Incident Detection Model | TP | FP | FN | Accuracy | Precision | Recall | F2-score |
|---|---|---|---|---|---|---|---|
| RF | 557.9 | 0.6 | 11.1 | 0.999 | 0.999 | 0.980 | 0.984 |
| SVM (RBF) | 545 | 0 | 24 | 0.999 | 1 | 0.958 | 0.966 |
| SVM (Poly, Degree=2) | 556 | 3 | 13 | 0.999 | 0.994 | 0.977 | 0.981 |
| XGBoost | 559 | 0 | 10 | 0.999 | 1 | 0.982 | 0.986 |
| NN | 553 | 11.6 | 9 | 0.999 | 0.979 | 0.974 | 0.977 |
| Hybrid (2 qubits) | 559 | 9.8 | 6 | 0.999 | 0.981 | 0.979 | 0.980 |
| Hybrid (4 qubits) | 563 | 8.3 | 6 | 0.999 | 0.984 | 0.989 | 0.987 |

**Results for DS-2**
  Since the models perform well when there is sufficient training data, we test the models for extreme cases when there is insufficient training data. We have created another dataset named DS-2, where the train-test split has been changed. There are only 15000 rows in the training set and 55000 rows in the test set. The results for this case are given in Table 2. Here we see a significant change in the performance of the baseline models. The hybrid model with 4-qubits shows the most robustness by achieving an F2-score of 0.985, which is the highest among all the models. The hybrid model with two qubits also performs well,

*Khan et al.*with an F2-score of 0.979. The XGBoost model performs well in terms of precision, but recall is more important in our study, so in terms of F2-score, it has been compromised. On average, the hybrid model with four qubits gets 936 true positives, 16 false negatives, and nine false positives. The lowest number of false negatives is the most important achievement of this model. Overall, the hybrid model achieves an average recall of 98.3% and an F2-score of 98.5%.

**TABLE 2 Comparison of Model Performance for DS-2**

| Incident Detection Model | TP | FP | FN | Accuracy | Precision | Recall | F2-score |
|---|---|---|---|---|---|---|---|
| RF | 913.2 | 2.4 | 38.8 | 0.999 | 0.997 | 0.959 | 0.966 |
| SVM (RBF) | 888 | 1 | 64 | 0.998 | 0.998 | 0.932 | 0.945 |
| SVM (Poly, Degree=2) | 907 | 9 | 45 | 0.999 | 0.990 | 0.953 | 0.959 |
| XGBoost | 926 | 4 | 26 | 0.999 | 0.996 | 0.973 | 0.977 |
| NN | 906.8 | 4.4 | 45.2 | 0.999 | 0.995 | 0.953 | 0.961 |
| Hybrid (2 qubits) | 930.4 | 8.8 | 21.6 | 0.999 | 0.991 | 0.977 | 0.979 |
| Hybrid (4 qubits) | 935.8 | 8.6 | 16.2 | 0.999 | 0.991 | 0.983 | 0.985 |

**Results for DS-3**

We have done further aggregation and aggregated the dataset by time, resulting in a per-minute dataset compared to a per-second dataset previously. This dataset is DS-3. This is the most extreme case of incident detection. Here, the number of rows has reduced significantly, and the number of positive labels is very scarce. Moreover, we have split 150 rows into training and 1250 rows into testing, so the models have very scarce data with a low number of positive labels to learn the incident detection. In this case, we see even more differences between the hybrid model and other baseline models. The SVM (RBF) and XGBoost model is unable to learn anything from the dataset. Hence they have recalls close to 0. The RF model has 14.1 average true positives but suffers from false negatives and lower recall. The SVM model with 2$^{nd}$ degree polynomial kernel predicts all 18 true positives but suffers from very low precisions since there are many false positives. The classical neural network also performs similarly. However, 2-qubit and 4-qubit hybrid models perform significantly better than all other models. The hybrid model with four qubits has an average false positive of 1.1 and an average false negative of 0.6. These results are very promising since we are dealing with extreme cases where there is insufficient data to train regular machine learning models. The hybrid model with four qubits achieves an accuracy of 0.998 and an F2-score of 0.961, which shows that hybrid classical-quantum neural networks are very suitable for incident detection when there is a lack of data. Table 3 shows the result from this analysis.

**TABLE 3 Comparison of Model Performance for DS-3**

| Incident Detection Model | TP | FP | FN | Accuracy | Precision | Recall | F2-score |
|---|---|---|---|---|---|---|---|
| RF | 14.1 | 0 | 3.9 | 0.996 | 1 | 0.783 | 0.817 |
| SVM (RBF) | 1 | 0 | 17 | 0.984 | 1 | 0.055 | 0.068 |
| SVM (Poly, Degree=2) | 18 | 11 | 0 | 0.989 | 0.621 | 1 | 0.891 |
| XGBoost | 0 | 0 | 18 | 0.983 | Nan | 0 | Nan |
| NN | 18 | 12.1 | 0 | 0.988 | 0.599 | 1 | 0.881 |
| Hybrid (2 qubits) | 17.4 | 1.8 | 0.6 | 0.997 | 0.908 | 0.966 | 0.953 |
| Hybrid (4 qubits) | 17.4 | 1.1 | 0.6 | 0.998 | 0.941 | 0.966 | 0.961 |

From the results, it can be concluded that hybrid classical-quantum neural networks offer two major advantages over classical neural networks and other machine learning models. First, hybrid models can learn from scarce training data and achieve higher accuracy for test cases than other models. Second, hybrid neural networks need a lower number of epochs to reach high accuracy compared to classical neural networks, which means that the models can be trained faster.

**CONCLUSIONS**

Quantum computing can be a promising avenue for traffic incident detection. This study showed that the hybrid neural network that contains both classical and quantum layers performs better when there is a lack of training data available for incidents on the road. This can be very useful in the real world since incident data is very scarce, and there is a lack of sufficient data for detecting future incidents.

Also, the hybrid model has fewer number of epochs, which correspond to a faster training process. This is useful for model updates and real-time applications. Overall, the hybrid model shows more robustness to different incident detection challenges, and it should be investigated more for improving roadway traffic safety in the future. With the continuing improvements of quantum computing infrastructure and algorithms, the quantum machine learning models could be a promising alternative for connected vehicle-related applications when the available data is insufficient.

**ACKNOWLEDGMENTS**

This study is supported by the Center for Connected Multimodal Mobility (C2M2) (a US Department of Transportation Tier 1 University Transportation Center) headquartered at Clemson University, Clemson, South Carolina, USA. Any opinions, findings, conclusions, and recommendations expressed in this material are those of the author(s). They do not necessarily reflect the views of C2M2, and the US Government assumes no liability for the contents or use thereof.

**AUTHOR CONTRIBUTIONS**

The authors confirm contribution to the paper as follows: study conception and design: Z. Khan, S.M. Khan, Jean, Ayse, Diamon, M. Chowdhury, R. Majumdar, J. Mwakalonge, D. Michalaka, G. Comert; data collection: Z. Khan, Jean, S.M. Khan; analysis and interpretation of results: Z. Khan, S.M. Khan, M. Chowdhury, Jean; draft manuscript preparation: Z. Khan, Jean, Ayse, Diamon, S.M. Khan, M. Chowdhury, D. Michalaka, J. Mwakalonge, G. Comert. All authors reviewed the results and approved the final version of the manuscript.